\documentclass[runningheads]{llncs}
\usepackage[utf8]{inputenc} % Swedish letters
\usepackage{graphicx}
\usepackage{amsmath}
\usepackage{subfigure}
\usepackage{caption}

\begin{document}
\title{Generating diffusion MRI scalar maps from T1 weighted images using generative adversarial networks}
\titlerunning{Generating diffusion MRI scalar maps using GANs}
\author{Xuan Gu\inst{1,2} \and
Hans Knutsson\inst{1,2} \and
Markus Nilsson\inst{4} \and
Anders Eklund\inst{1,2,3}}

\authorrunning{X. Gu et al.}

\institute{Division of Medical Informatics, Department of Biomedical Engineering \and
Center for Medical Image Science and Visualization (CMIV) \and
Division of Statistics and Machine learning, Department of Computer and Information Science \\
Linköping University, Linköping, Sweden\and
Department of Clinical Sciences, Radiology, Lund University, Lund, Sweden\\
\email{xuan.gu@liu.se}}

\maketitle              % typeset the header of the contribution
\begin{abstract}
Diffusion magnetic resonance imaging (diffusion MRI) is a non-invasive microstructure assessment technique. Scalar measures, such as FA (fractional anisotropy) and MD (mean diffusivity), quantifying micro-structural tissue properties can be obtained using diffusion models and data processing pipelines. However, it is costly and time consuming to collect high quality diffusion data. Here, we therefore demonstrate how Generative Adversarial Networks (GANs) can be used to generate synthetic diffusion scalar measures from structural T1-weighted images in a single optimized step. Specifically, we train the popular CycleGAN model to learn to map a T1 image to FA or MD, and vice versa. As an application, we show that synthetic FA images can be used as a target for non-linear registration, to correct for geometric distortions common in diffusion MRI.
\end{abstract}

\section{Introduction}

Diffusion MRI is a non-invasive technique used for studying brain tissue microstructures. Diffusion-derived scalar maps provide rich information about microstructural characterization, but there are two major bottlenecks in obtaining these scalar maps. First, it is expensive and time consuming to acquire high quality diffusion data. Second, the accuracy of the diffusion-derived scalar maps relies on elaborate diffusion data processing pipelines, including preprocessing (head motion, eddy current distortion and susceptibility-induced distortion corrections), diffusion model fitting and diffusion scalar calculation. Small errors occurring at any of these steps can contribute to the bias of the diffusion-derived scalars. For some advanced diffusion models e.g. mean apparent propagator (MAP) MRI \cite{ozarslan2013mean}, processing of a single slice of the brain can take hours to finish.

Generative Adversarial Networks (GANs) is one of the most important ideas in machine learning in the last 20 years \cite{goodfellow2014generative}. GANs have already been widely used for medical image processing applications, such as denoising, reconstruction, segmentation, detection, classification and image synthesis. However, GANs for medical image translation are still rather unexplored, especially for cross-modality translation of MR images \cite{kazeminia2018gans}.

Implementations of CycleGAN and unsupervised image-to-image translation (UNIT) for 2D T1-T2 translation were reported in \cite{welander2018generative}, and results showed that visually realistic synthetic T1 and T2 images can be generated from the other modality, proven via a perceptual study.
In \cite{dar2018image} a conditional GAN  was proposed to do the translation between T1 and T2 images, in which a probabilistic GAN (PGAN) and a CycleGAN were trained for paired and unpaired source-target images, respectively. The proposed GAN demonstrated visually and quantitatively accurate translations for both healthy subjects and glioma patients.
A patch-based conditional GAN \cite{olut2018generative} was proposed to generate magnetic resonance angiography (MRA) images from T1 and T2 images jointly. Steerable filter responses were incorporated in the loss function to extract directional features of the vessel structure.
MR image translation based on downsampled images was investigated in \cite{dar2018synergistic}, to reduce scan time. Three different types of input were fed to the GANs: downsampled target images, downsampled source images, and downsampled target and source images jointly. It was demonstrated that the GAN with the combination of downsampled target and source images as the input outperformed its two competitors in reconstructing higher resolution images, which resulted in a reduction of the scan time up to a factor of 50.
A 3D conditional GAN \cite{yu20183d} was applied to synthesize FLAIR images from T1, and the synthetic FLAIR images  improved brain tumor segmentation, compared with using only T1 images.

In this work, we explored the possibility to generate diffusion scalar maps from structural MR images. We propose a new application of CycleGAN \cite{zhu2017unpaired}; to translate T1 images to diffusion-derived scalar maps. To the best of our knowledge, this is the first study of GAN-based MR image translation between structural space and diffusion space. Both qualitative and quantitative evaluations of generated images were carried out in order to assess effectiveness of the method. We also show how synthetic FA images can be used as a target for non-linear registration, to correct for geometric distortions common in diffusion MRI.

\section{Theory}
\subsection{Diffusion tensor model}

In a diffusion experiment, the diffusion-weighted signal $S_i$ of the $i$th measurement for one voxel  is modeled by
\begin{align}
  S_i = S_0 \exp(-b \mathbf{g}_i^T \mathbf{Dg}_i), \quad \mathrm{for} \quad i = 1, 2, \cdots, T,
\end{align}
where $S_0$ is the signal without diffusion weighting, $b$ is the diffusion weighting factor, $\mathbf{D}=\begin{bmatrix}
  D_{xx} & D_{xy} & D_{xz} \\
  D_{xy} & D_{yy} & D_{yz} \\
  D_{xz} & D_{yz} & D_{zz}
\end{bmatrix}$ is the diffusion tensor in the form of a $3 \times 3$ positive definite matrix, $\mathbf{g}_i$ is a $3 \times 1$ unit vector of the gradient direction, and $T$ is the total number of measurements. Mean diffusivity (MD) and fractional anisotropy (FA) can be calculated from the estimated tensor, according to

\begin{align}
  MD&=(\lambda_1+\lambda_2+\lambda_3)/3, \\
  FA&=\sqrt{\frac{(\lambda_1-\lambda_2)^2+(\lambda_2-\lambda_3)^2+(\lambda_3-\lambda_1)^2}{2(\lambda_1^2+\lambda_2^2+\lambda_3^2)}},
\end{align}
where $\lambda_1$, $\lambda_2$ and $\lambda_3$ are the three eigenvalues of the diffusion tensor $\mathbf{D}$.
In our case weighted least squares was used to estimate the diffusion tensor.

\subsection{CycleGAN}
A CycleGAN \cite{zhu2017unpaired} can be trained using two unpaired groups of images, to translate images between domain A and domain B. A CycleGAN consists of four main components, two generators ($G_{A2B}$ and $G_{B2A}$) and two discriminators ($D_A$ and $D_B$). The two generators synthesize domain A/B images based on domain B/A. The two discriminators are making the judgement if the input images belong to domain A/B. The translation between the two image domains is guaranteed by
\begin{align}
  G_{B2A}(G_{A2B}(I_A)) &\approx I_A \\
  G_{A2B}(G_{B2A}(I_B)) &\approx I_B
  \label{eq:cycle}
\end{align}
where $I_A$ and $I_B$ are two images of domain A and B.
The loss function contains two terms: adversarial loss and cycle loss, and can be written as \cite{zhu2017unpaired}
\begin{align}
  L_{adv}&=E_{a \in A}[(D_A(a)-1)^2]+E_{b \in B}[(D_B(b)-1)^2]\\
  &+E_{a \in A}[D_B(G_{A2B}(a))^2]+E_{b \in B}[D_A(G_{B2A}(b))^2]\\
  &+E_{a \in A}[(D_B(G_{A2B}(a))-1)^2]+E_{b \in B}[ (D_A(G_{B2A}(b))-1)^2],\\
  L_{cyc}&=E_{a \in A}[|G_{B2A}(G_{A2B}(a))-a|] + E_{b \in B}[|G_{A2B}(G_{B2A}(b))-b|].
\end{align}
The adversarial loss encourages the discriminators to approve the images of the corresponding groups, and reject the images that are generated by the corresponding generators. The generators are also encouraged to generate images that can fool the corresponding discriminators.
The cycle loss guarantees that the image can be reconstructed from the other domain, as stated in Equation \ref{eq:cycle}.
The total loss is the sum of the adversarial loss and the cycle loss, i.e.
\begin{align}
  L_{total} = L_{adv} + L_{cyc}.
\end{align}

\subsection{Similarity measure}
The widely used structural similarity (SSIM) measure~\cite{wolterink2017deep,emami2018generating} was used to quantify the accuracy of the image translations. SSIM can measure local structural similarity between two images.
The SSIM quantifies the degree of similarity of two images based on the impact of three characteristics: luminance, contrast and structure. The SSIM of pixel $(x,y)$ in images A and B can be calculated as \cite{wang2004image}
\begin{align}
  SSIM(x,y) = \frac{(2 \mu_{w_A} \mu_{w_B} +c_1)(2 \sigma_{w_Aw_B}+c_2)}{(\mu_{w_A}^2+\mu_{w_B}^2+c1)(\sigma_{w_A}^2+\sigma_{w_B}^2+c_2)},
\end{align}
where $w_A$ and $w_B$ are local neighborhoods centered at $(x,y)$ in images A and B, $\mu_{w_A}$ and $\mu_{w_B}$ are the local means, $\sigma_{w_A}$ and $\sigma_{w_B}$ are the local standard deviations, $\sigma_{w_Aw_B}$ is the covariance, $c_1$ and $c_2$ are two variables to stabilize the division. The mean SSIM ($MSSIM=SSIM/N_{voxel}$) within the brain area can be used as a global measure of the similarity between the synthetic image and the ground truth.

\section{Data}
\label{sec:pagestyle}
We used diffusion and T1 images from the Human Connectome Project (HCP)\footnote{Data collection and sharing for this project was provided by the Human Connectome Project (U01-MH93765) (HCP; Principal Investigators: Bruce Rosen, M.D., Ph.D., Arthur W. Toga, Ph.D., Van J.Weeden, MD). HCP funding was provided by the National Institute of Dental and Craniofacial Research (NIDCR), the National Institute of Mental Health (NIMH), and the National Institute of Neurological Disorders and Stroke (NINDS). HCP data are disseminated by the Laboratory of Neuro Imaging at the University of Southern California.} \cite{van2013wu,glasser2013minimal}  for 1065 healthy subjects. The data were collected using a customized Siemens 3T Connectom scanner. The diffusion data were acquired with 3 different b-values (1000, 2000, and 3000 s/mm$^2$) and have already been pre-processed for gradient nonlinearity correction, motion correction and eddy current correction. The diffusion data consist of 18 non-diffusion weighted volumes (b = 0) and 90 volumes for each b-value, which yields 288 volumes of $145 \times 174 \times 145$ voxels with an 1.25 mm isotropic voxel size. The T1 data was acquired with a $320 \times 320$ matrix size and $0.7 \times 0.7 \times 0.7$ mm isotropic voxel size, and then downsampled by us to the same resolution as the diffusion data.

\section{Methods}

Diffusion tensor fitting and MAP-MRI fitting were implemented using C++ and the code is available on Github\footnote{https://github.com/xuagu37/dtb}. We used a Keras implementation of 2D CycleGAN, originating from the work by~\cite{welander2018generative}, which is also available on Github\footnote{https://github.com/xuagu37/CycleGAN}.
The statistics analysis was performed in MATLAB (R2018a,
The MathWorks, Inc., Natick, Massachusetts, United
States).

We followed the network architecture design given in the original CyceGAN paper~\cite{zhu2017unpaired}. We used 2 feature extraction convolutional layers, 9 residual blocks and 3 deconvolutional layers for the generators. For the discriminators we used 4 feature extraction convolutional layers and a final layer to produce a one-dimensional output. We trained the network with a learning rate of 0.0004. We kept the same learning rate for the first half of the training, and linearly decayed the learning rate to zero over the second half. A total of 1000 subjects were used for training, and 65 subjects were used for testing. An Nvidia Titan X Pascal graphics card was used to train the network.
The experiment protocols and training times are summarized in Table \ref{experiments}.

 \begin{table}[]
   \centering
 \begin{tabular}{|c|c|c|c|}
 \hline
              & \begin{tabular}[c]{@{}c@{}}Training data\end{tabular}                       & \begin{tabular}[c]{@{}c@{}}Test data\end{tabular}                         & \begin{tabular}[c]{@{}c@{}}Training time (h)\end{tabular} \\ \hline
 Experiment 1 & \begin{tabular}[c]{@{}c@{}}1000 subjects\\ 1 slice per subject\end{tabular}   & \begin{tabular}[c]{@{}c@{}}65 subjects\\ 1 slice per subject\end{tabular}   & 6                                                           \\ \hline
 Experiment 2 & \begin{tabular}[c]{@{}c@{}}1000 subjects\\ 1 slice per subject\end{tabular}   & \begin{tabular}[c]{@{}c@{}}65 subjects\\ 17 slices per subject\end{tabular} & 6                                                           \\ \hline
 Experiment 3 & \begin{tabular}[c]{@{}c@{}}1000 subjects\\ 17 slices per subject\end{tabular} & \begin{tabular}[c]{@{}c@{}}65 subjects\\ 17 slices per subject\end{tabular} & 107                                                         \\ \hline
 \end{tabular}
 \vspace{0.2cm}
 \caption{Experiment protocols used to train and evaluate the CycleGAN}
 \label{experiments}
 \end{table}

\section{Results}
\subsection{Synthetic FA and MD}

\begin{figure}
\centering     %%% not \center
\subfigure[T1-to-FA.]{\label{fig:1a}\includegraphics[width=0.8\textwidth]{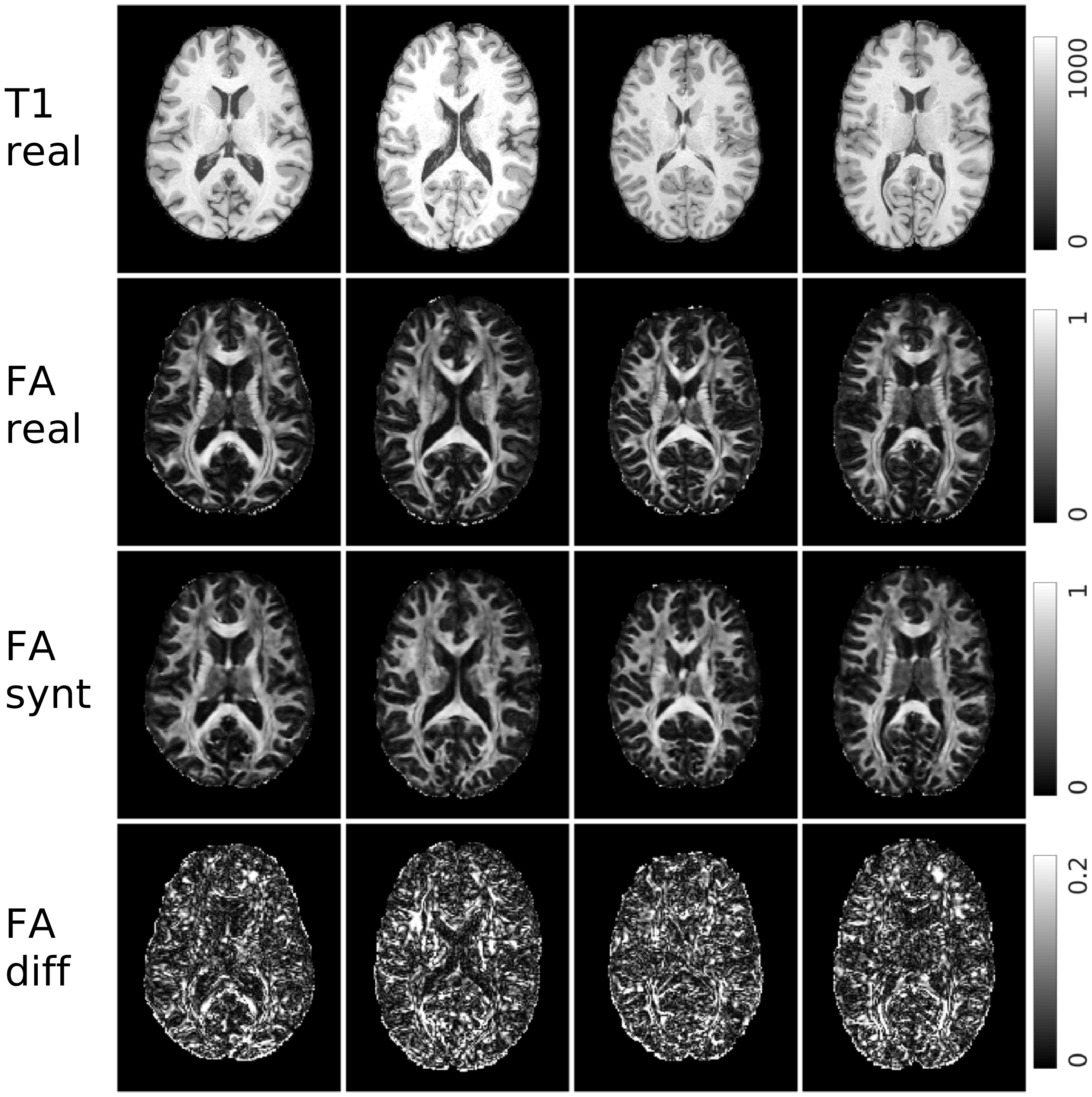}}
\subfigure[T1-to-MD.]{\label{fig:1b}\includegraphics[width=0.8\textwidth]{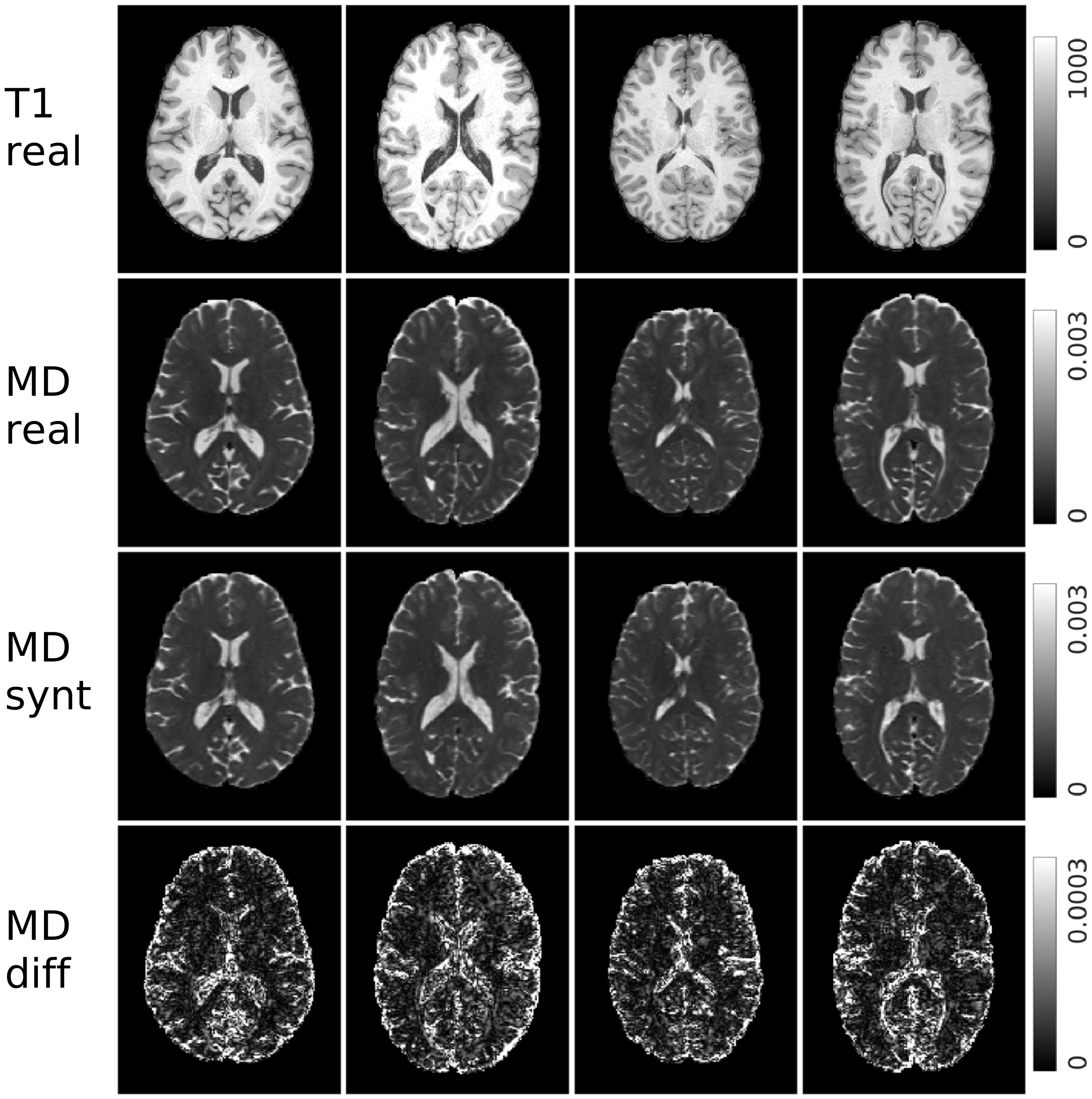}}
\caption{T1-to-FA/MD image translation results for 4 subjects. First row: True T1 images, second row: true FA images, third row: synthetic FA images, fourth row: Difference of true and synthetic FA.}
\end{figure}

Figure \ref{fig:1a} shows the qualitative results of T1-to-FA image translation for 4 test subjects. The results show a good match between the synthetic FA images and their ground truth, for both texture of white matter tracts and global content.
However, when compared with the ground truth, it is observed that the synthetic FA images have a reduced level of  details on white matter tracts. The difference image shows the absolute error between synthetic and real FA images.
Results of T1-to-MD image translation are shown in Figure \ref{fig:1b}. The synthetic MD images demonstrate great visual similarity to the ground truth. The CSF region and its boundaries are accurately synthesized.

\begin{figure}
\centering     %%% not \center
\subfigure[Experiment 1.]{\label{fig:2a}\includegraphics[width=0.55\textwidth]{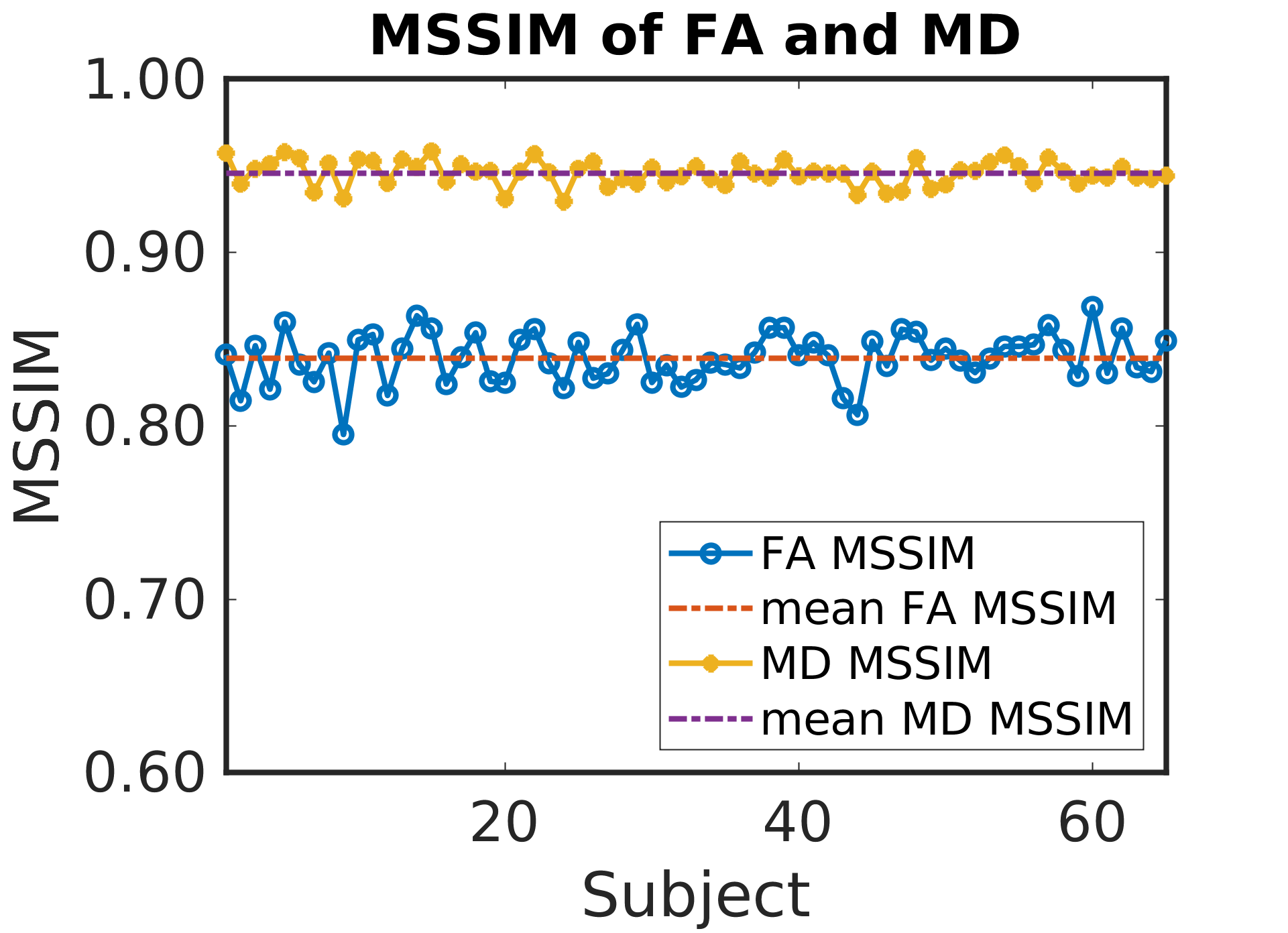}}
\subfigure[Experiment 2.]{\label{fig:2b}\includegraphics[width=0.55\textwidth]{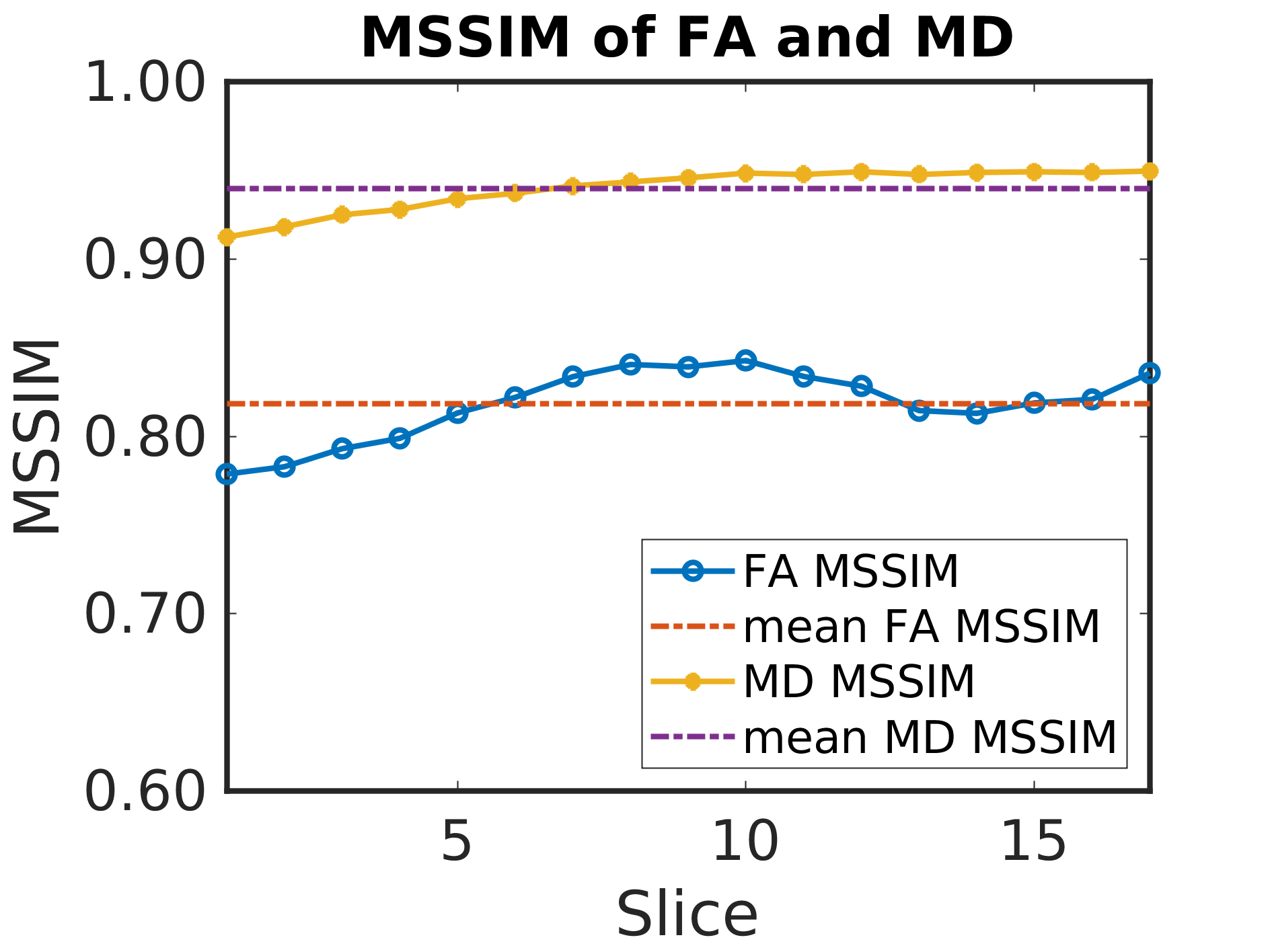}}
\subfigure[Experiment 3.]{\label{fig:2c}\includegraphics[width=0.55\textwidth]{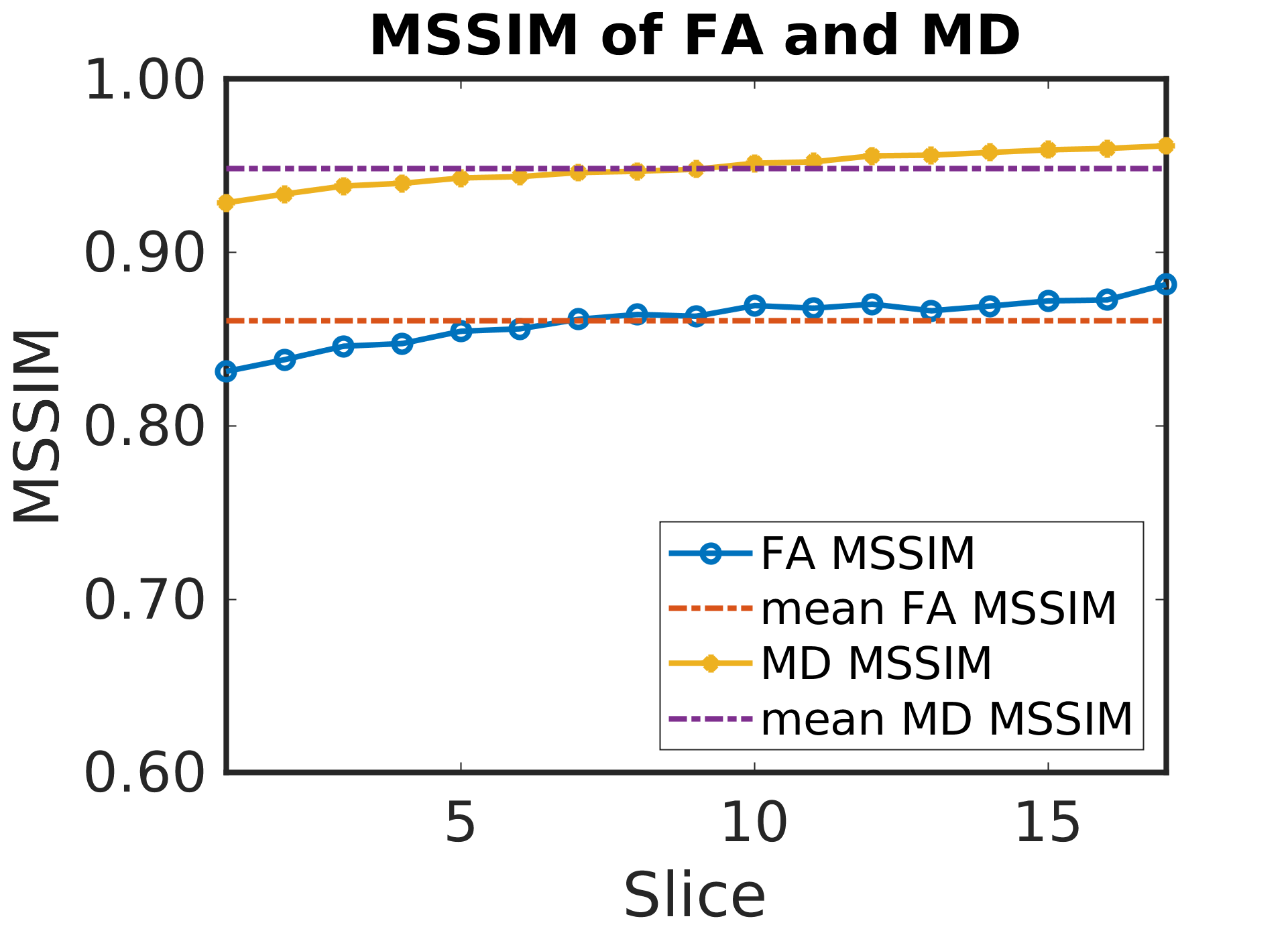}}
\caption{MSSIM results of FA and MD for Experiment 1, 2 and 3. Experiment 1: training on 1000 subjects, 1 slice per subject, and test on 65 subjects, 1 slice per subject. The mean MSSIM of synthetic FA and MD across subjects are 0.839 and 0.937, respectively. Experiment 2: training on 1000 subjects, 1 slice per subject, and test on 65 subjects, 17 slices per subject. The mean MSSIM of synthetic FA and MD across slices are 0.818 and 0.940, respectively. Experiment 3: training on 1000 subjects, 17 slices per subject, and test on 65 subjects, 17 slices per subject. The mean MSSIM of synthetic FA and MD across slices are 0.861 and 0.948, respectively. Plots show that synthetic MD images demonstrate higher accuracy compared to synthetic FA, and that using a higher number of training slices mostly helps FA synthesis.}
\label{fig:2}
\end{figure}

Figure \ref{fig:2a} shows the MSSIM of synthetic FA and MD images for the 65 test subjects. MSSIM values showed high consistency among the different test subjects. The mean$\pm$std intervals of the MSSIM are $0.839 \pm 0.014$ and $0.937 \pm 0.008$ for synthetic FA and MD images, respectively. The MSSIM results for synthetic MD are higher compared to synthetic FA. This may be partly due to that FA images contain richer structure information, thus it is more difficult to synthesize (since FA is more non-linear than MD). Figure \ref{fig:2b} and \ref{fig:2c} show the MSSIM of synthetic FA and MD images for the 17 slices. It can been that the MSSIM result is sensitive to the slice position, and that a higher number of training slices leads to higher MSSIM results for the synthetic FA and MD maps.

\subsection{Non-linear registration for distortion correction}

\begin{figure}
\centering     %%% not \center
\subfigure[FA LR.]{\label{fig:3a}\includegraphics[width=0.32\textwidth]{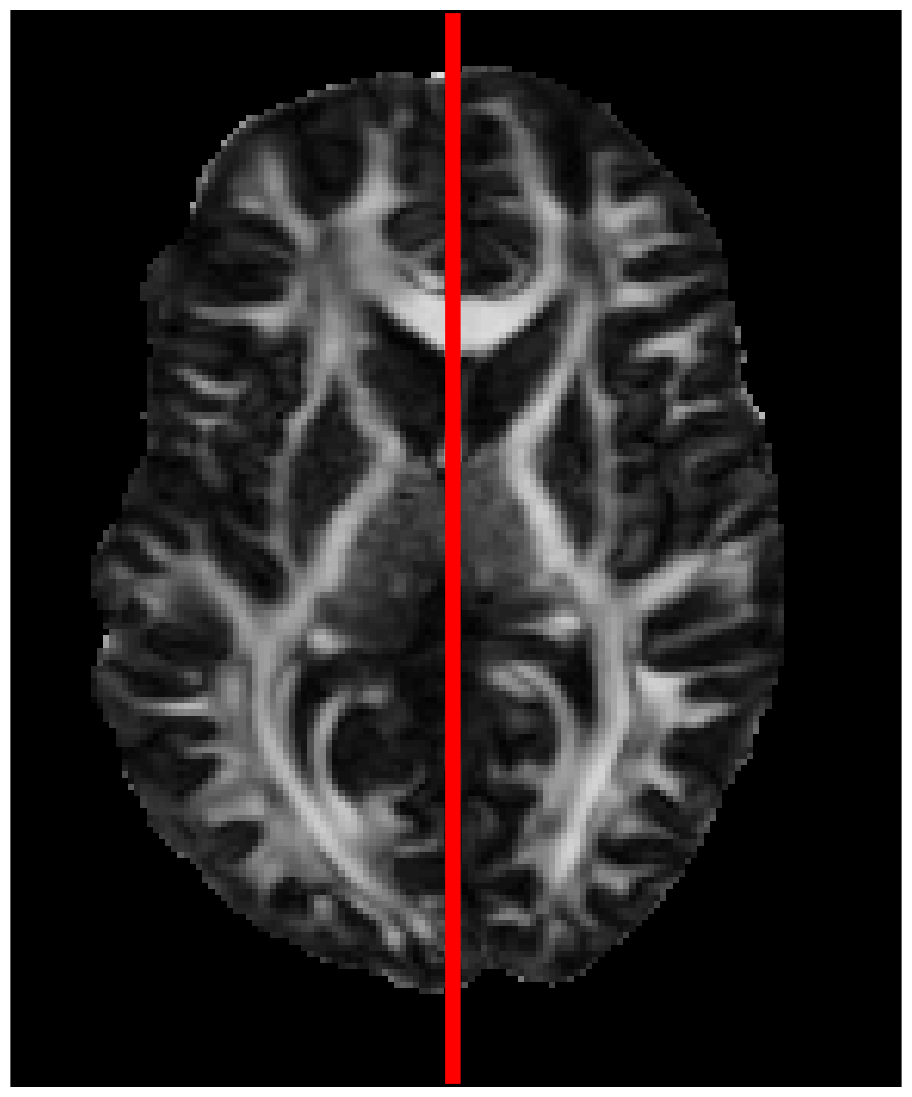}}
\subfigure[FA RL.]{\label{fig:3b}\includegraphics[width=0.32\textwidth]{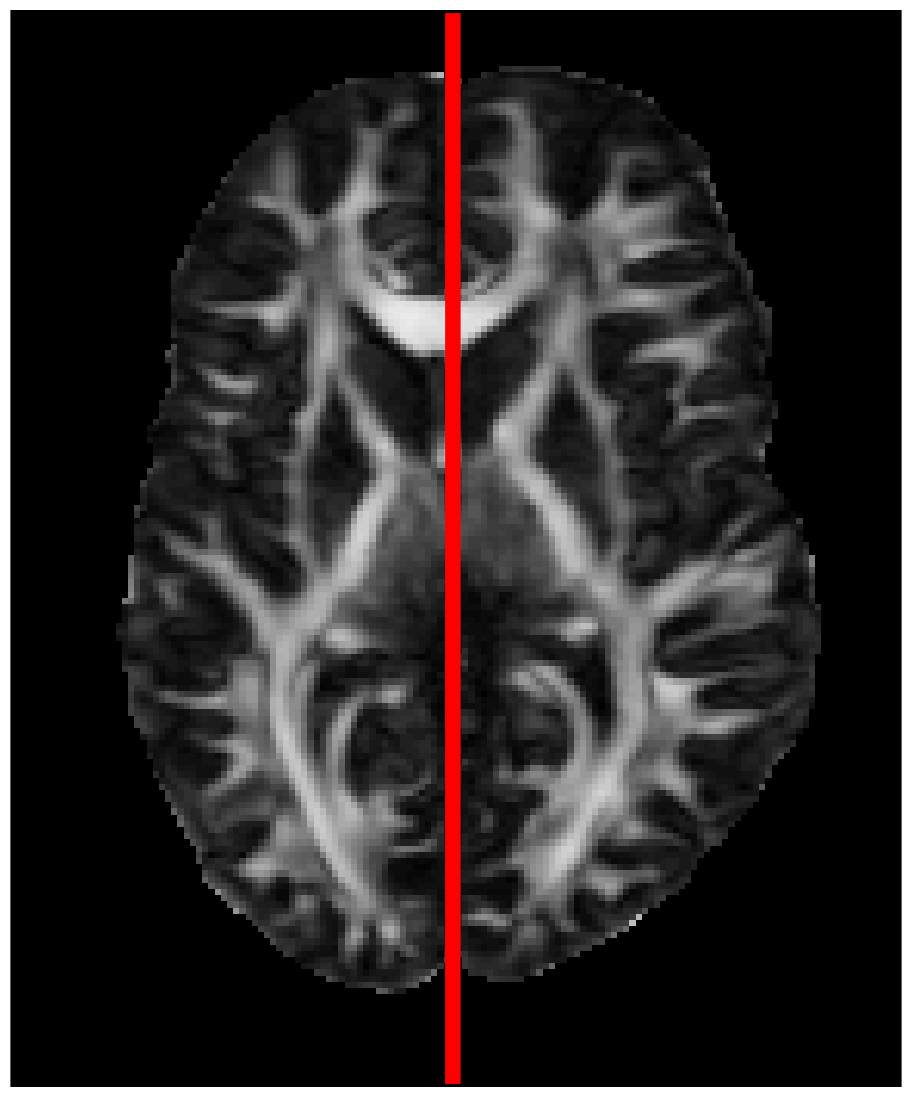}}
\subfigure[FA synthetic.]{\label{fig:3c}\includegraphics[width=0.32\textwidth]{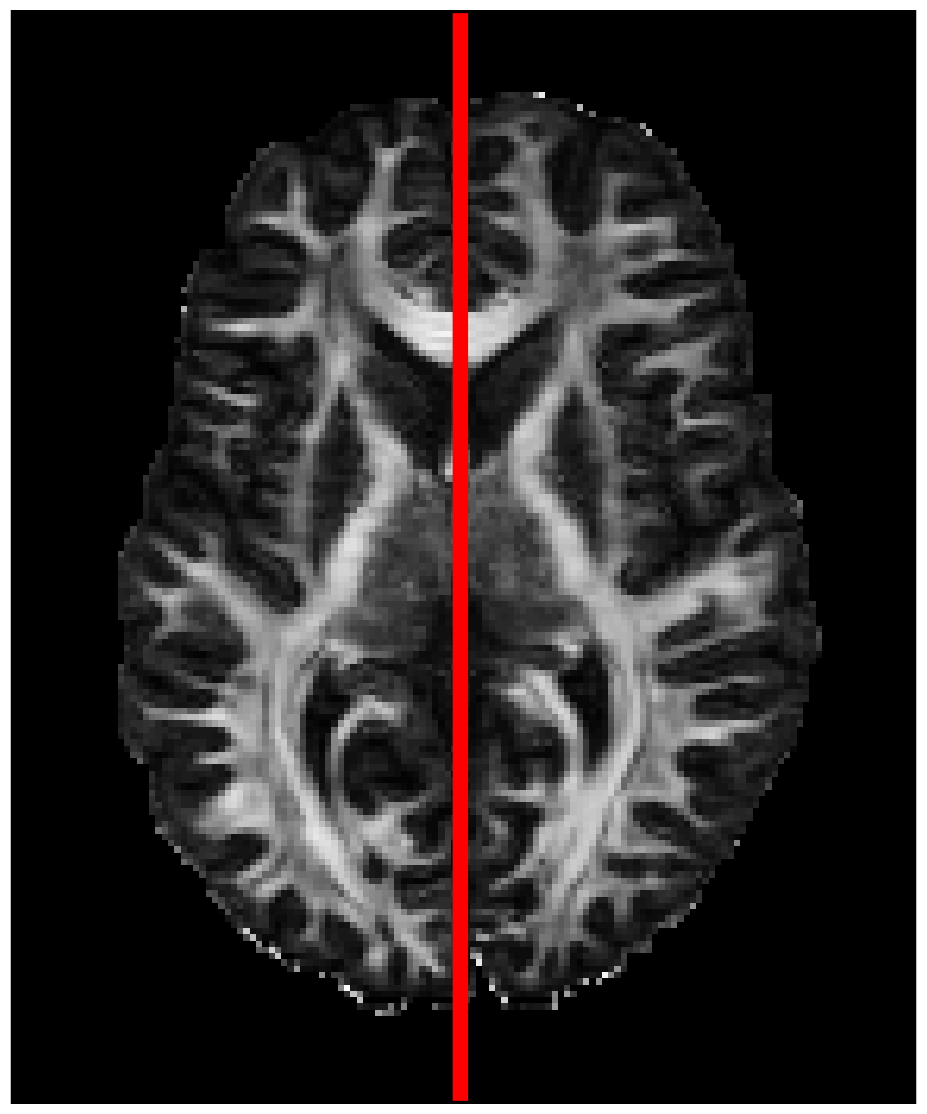}}
\subfigure[FA LR registered.]{\label{fig:3d}\includegraphics[width=0.32\textwidth]{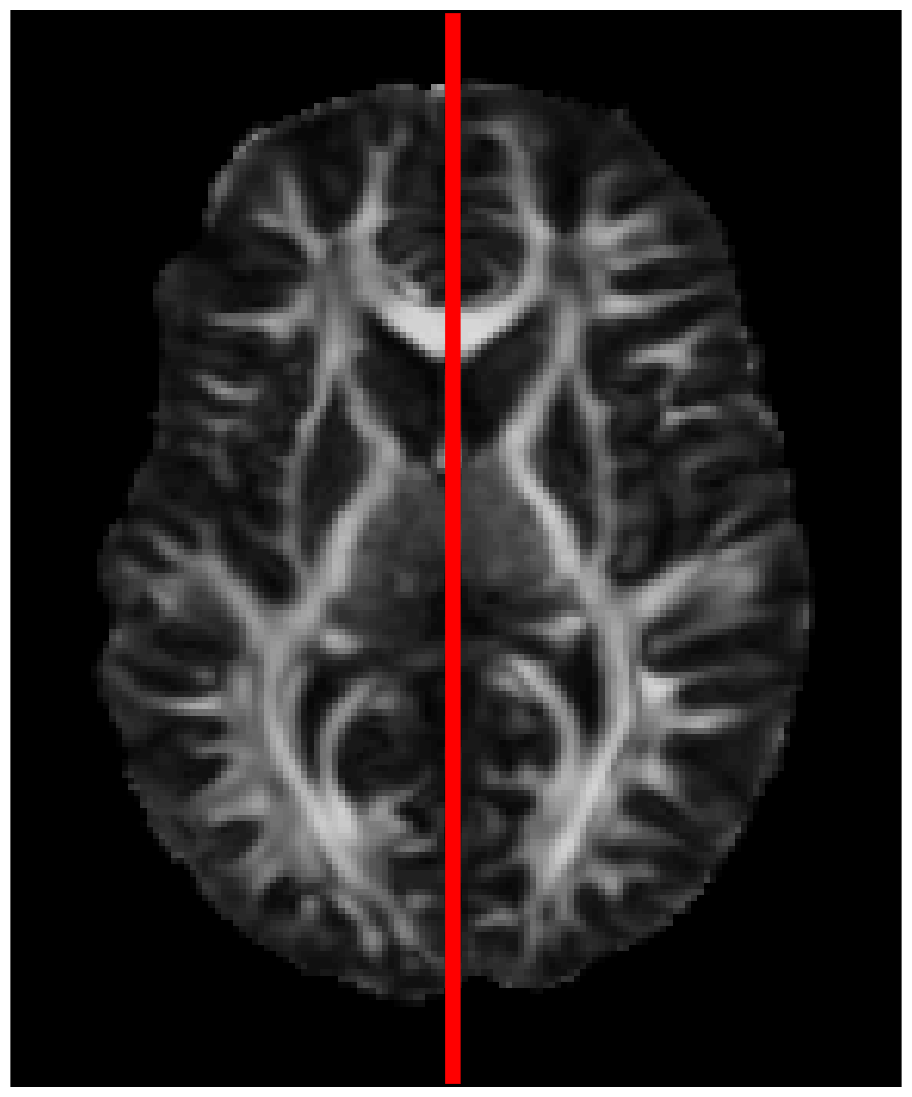}}
\subfigure[FA topup.]{\label{fig:3e}\includegraphics[width=0.32\textwidth]{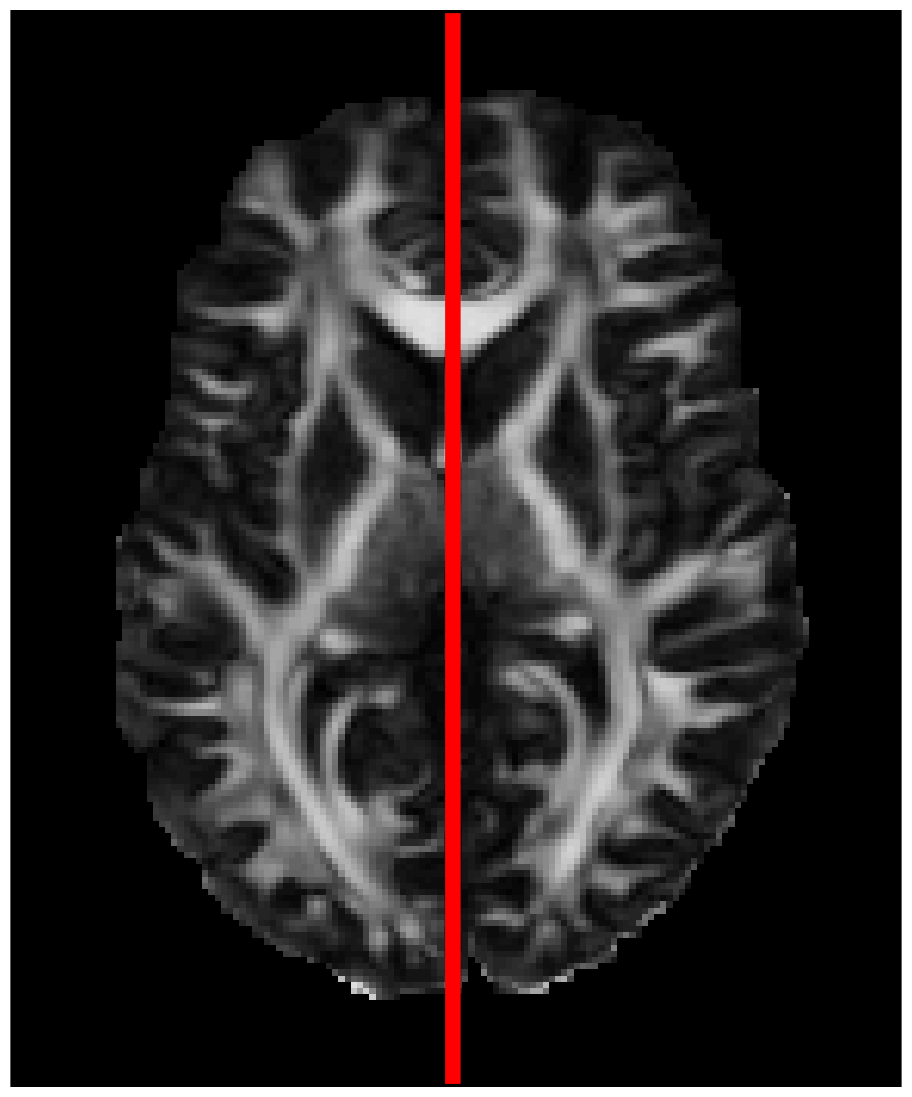}}
\caption{FA LR: FA map from data with left to right phase encoding direction. FA RL: FA map from data with right to left phase encoding direction. FA synthetic: FA map from CycleGAN. FA LR registered: FA LR non-linearly registered to synthetic FA. FA topup: FA map from EPI distortion corrected data (seen as gold standard). The benefit of using non-linear registration for distortion correction, instead of topup, is that the scan time is reduced by a factor 2.}
\label{fig:3}
\end{figure}

EPI distortions can be corrected by the FSL function topup, for data acquired with at least two phase encoding directions. However, it is hard to correct for EPI distortions for data acquired with a single phase encoding direction. A potential approach would be to generate a synthetic FA map from the undistorted T1 image, and then (non-linearly) register the distorted FA map to the undistorted synthetic one. This transform can then be applied to all other diffusion scalar maps. The FA map from EPI distortion corrected data (using topup) can be regarded as the gold standard. We used FNIRT in FSL to perform the non-linear registration. Figure \ref{fig:3} shows various FA maps for one test subject. The FA map from the proposed approach provides a result which is very similar to the gold standard. The benefit of our approach is that the scan time can be reduced a factor 2, by acquiring data using a single phase encoding direction. It is of course theoretically possible to register the distorted FA image directly to the undistorted T1 image, but non-linear registration of images with different intensity can be rather challenging.

\clearpage

\section{Discussion}

Translation between structural and diffusion images has been shown using CycleGAN. The synthetic FA and MD images are remarkably similar to their ground truth. Quantitative evaluation using MSSIM of 65 test subjects shows
that the trained CycleGAN works well for all test subjects, and that training using a larger number of slices improves the results.

While the synthetic FA images appear realistic, the training of the GAN will depend on the training data used. For example, if the GAN is trained using data from healthy controls it is likely that the GAN will be biased for brain tumor patients, and for example remove existing tumors \cite{cohen2018distribution}.

Future research may focus on creating other diffusion-derived scalar maps from more advanced diffusion models, such as mean apparent propagator (MAP) MRI. In this work, we have only used 2D CycleGAN, but it has been reported that 3D GANs using spatial information \cite{nie2017medical} across slices yield better mappings between two domains (at the cost of a higher memory consumption and a higher computational cost). A comparison study of image translation using 2D and 3D GANs is thus worth looking into.

%\section{Conclusion}
%Translation between MR structural and diffusion domains has been shown using Cycle-GAN. The synthetic FA and MD images
%are remarkably similar to their ground truth.
%Quantitative evaluation using MSSIM and CC of 65 test subjects shows
%that the trained Cycle-GAN works well for all test subjects.

%Future research may focus on creating other diffusion-derived scalar maps from more advanced diffusion models, such as mean apparent propagator (MAP) MRI. Also, we will investigate whether training using more than one slice per subject can further improve performance. In this work, we have only used 2D Cycle-GAN, but it has been reported that 3D GANs using spatial information \cite{nie2017medical} across slices yield better mappings between two domains. A comparison study of image translation using 2D and 3D GANs is thus worth looking into.

\section*{Acknowledgements}

This study was supported by Swedish research council grants 2015-05356 and 2017-04889. Funding was also provided by the Center for Industrial Information Technology (CENIIT) at Linköping University, the Knut and Alice Wallenberg foundation project ”Seeing organ function”, Analytic Imaging Diagnostics Arena (AIDA) and the ITEA3 / VINNOVA funded project "Intelligence based iMprovement of Personalized treatment And Clinical workflow supporT" (IMPACT). The Nvidia Corporation, who donated the Nvidia Titan X Pascal graphics card used to train the GANs, is also acknowledged.

\bibliographystyle{splncs04}
\bibliography{mybibliography}

\end{document}